\documentclass[letterpaper, 10 pt, conference]{ieeeconf}
\IEEEoverridecommandlockouts                              
\usepackage[utf8]{inputenc}
\usepackage{cite}
\usepackage{graphicx}
\usepackage{amsmath}
\usepackage{bm}
\usepackage{hyperref}
\usepackage{color}
\graphicspath{{figures/}}

\newcommand{\xv}{{\bf x}}
\newcommand{\dv}{{\bf d}}

\newcommand{\Rm}{{\bf R}}

\usepackage[left=1.72cm, right=1.72cm, top=2.12cm, bottom=1.72cm]{geometry}

\begin{document}

\title{Self Contained Relative Localization with a Low-Cost Multi-Robot System}
\author{I. D. Miller$^{1}$ and J. Wallace$^{2}$
%\thanks{*This work was supported in part by a Lafayette EXCEL grant.}
\thanks{$^{1}$Ian D. Miller was with the Dept. of Electrical and Computer Engineering, Lafayette College, Easton, PA, USA, and is now with the University of Pennsylvania GRASP Lab, Philadelphia, PA, USA.
        {\tt\small iandm@seas.upenn.edu}}
\thanks{$^{2}$Jon Wallace is with the Dept. of Electrical and Computer Engineering,
        Lafayette College, Easton, PA, USA.
        {\tt\small wallacjw@lafayette.edu}}
}

\maketitle

\begin{abstract}
    A key limitation of current multi-robot systems is a lack of relative localization, particularly in environments without GPS or motion capture systems.  This article presents a centralized method for relatively localizing a 2D swarm using sensors and beacons on the robots themselves.  The UKF-based algorithm as well as the requisite novel and cost-effective sensing hardware are discussed.  Comparisons with a motion capture system show that the method is capable of localization with errors on the order of the size of the robots.
\end{abstract}

\begin{keywords}
    Localization, Multi-Robot Systems, Range Sensing.
\end{keywords}

%-----------------------------------------
\section{Introduction}

\PARstart{S}{ince} its inception, robotics has been inspired by biology, and recent multi-robot systems can emulate the swarming behavior seen in the animal kingdom. In an early paper, Kube successfully built a simple ground-based swarm system using simple rules inspired by insects\cite{kube}.  Since then, swarms have taken to the sky \cite{mulgaonkar} and even the water \cite{waduge}.  Robot swarms are not only a fascinating achievement, but also have useful applications ranging from search and rescue \cite{kumaras} to satellite constellations \cite{gill}.  Swarms have key advantages over traditional single robot systems with respect to redundancy, resiliency, and parallelization \cite{brambilla}.

Knowledge of system state is a key requirement for most robot platforms, and in the case of a multi-robot system or swarm, the system state includes the relative locations and orientation of the individual members.  Centralized and real-time knowledge of this state is desirable for a human or autonomous controller.  A simple and often-used method for state estimation employs an external motion capture system.  This approach is often used for swarms such as the Robotarium \cite{pickem}, but such external systems often preclude practical deployment.

A conceptually straightforward approach to multi-robot localization is to individually globally localize each member, allowing any member or controller to easily compute relative locations with minimal communication.  Global localization can be performed with GPS or visual odometry as in \cite{weinstein}, but GPS is expensive, limited in accuracy, and cannot be used indoors, while odometry methods gradually drift and accrue error over time.

Given the difficulties of global localization, attention has focused on robot systems that estimate relative robot position using onboard sensors and combine this information to obtain the global system state.  In \cite{roumeliotis} and \cite{martinelli} the Kalman filter was employed, but the method was not tested on real hardware and no solution for obtaining the initial system state was given.  More recently, in \cite{prorok} a method using particle filters is described and tested on actual Khepera III robots, but the approach has significant computational demand.  Several sensing methods have been proposed to obtain robot bearing and range, including optical tags \cite{saska}, sonar \cite{boussetta}, and infrared (IR) \cite{prorok}, \cite{rubenstein}.  These strategies rely either on relatively expensive hardware or, in the case of \cite{rubenstein}, provide only bearing information.

Given this landscape, experimental swarm research is out of the reach of many interested researchers, either due to the high cost or non-deployable nature of the specialized hardware.  Therefore, the purpose of this paper is to present a robot platform employing a local sensing method that is deployable in real scenarios, yet has a very low cost.  Specifically, we present a system capable of obtaining real-time position information of swarm members with an accuracy better than the size of a robot, without requiring known initial conditions.  An external motion capture system is only used to study position estimation error and is not required in actual deployment.  Further, low cost was a critical objective, which was kept within \$200 per node. 

We note similarity to the work in \cite{martinelli}, but unlike the distributed sensing algorithm there that requires inter-robot communication, we limit our present scope to centralized computation, where nodes report sensor data directly to a controller.  We also introduce a novel low-cost sensor system for gathering the requisite relative pose information.  Details of the hardware and software components developed in this project are available at no cost to interested researchers
\footnote{https://github.com/iandouglas96/swarm-thesis/ under the MIT license.}.

%-----------------------------------------
\section{Robotic Platform}
\label{section:platform}
\begin{figure}
  \centering
  \includegraphics[width=0.5\textwidth]{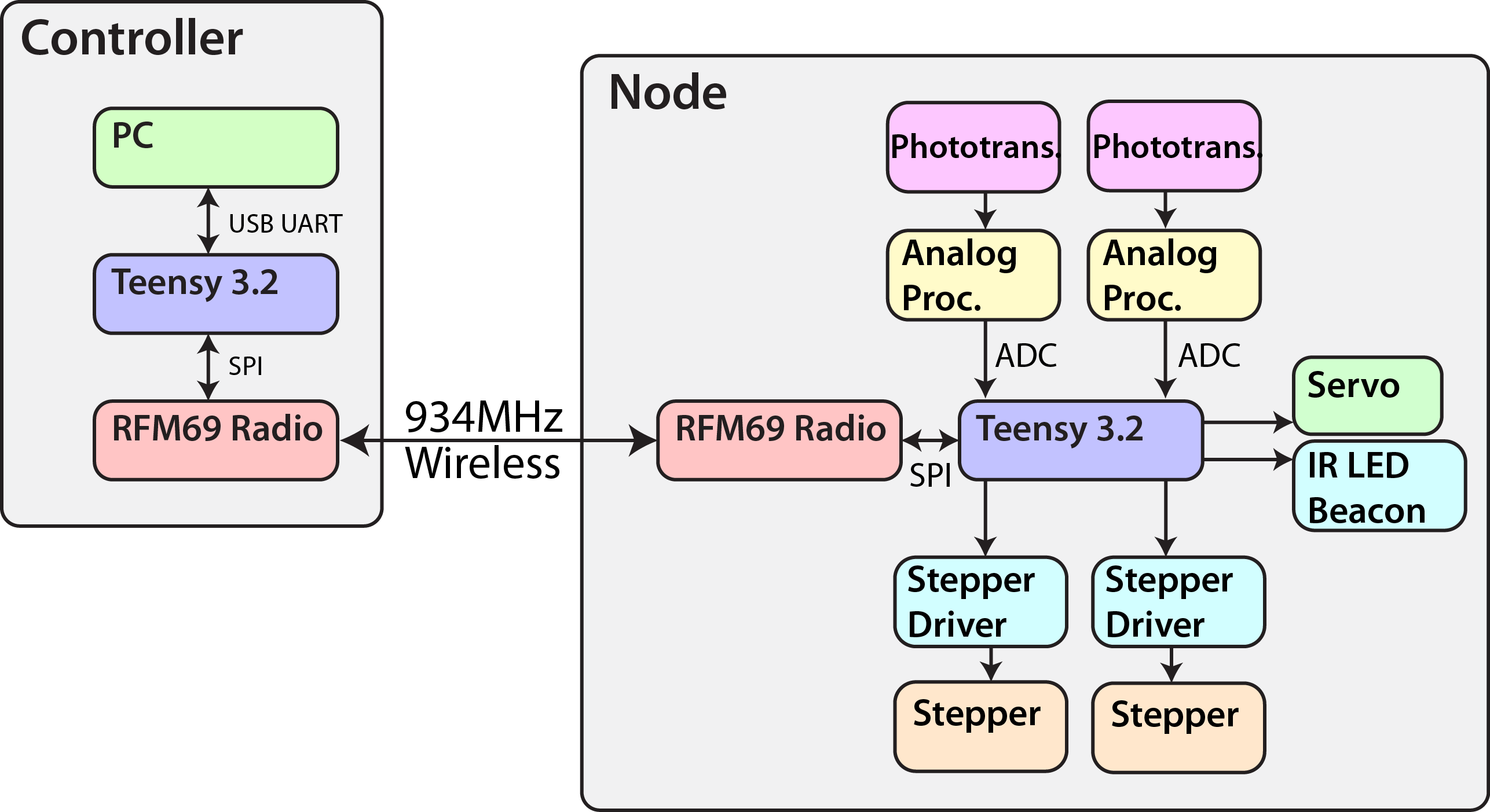}
  \caption{High-level hardware block diagram of swarm robot and     central controller.}
  \label{fig:hardware_block_diag}
\end{figure}

\begin{figure}
    \centering
    \includegraphics[width=0.3\textwidth]{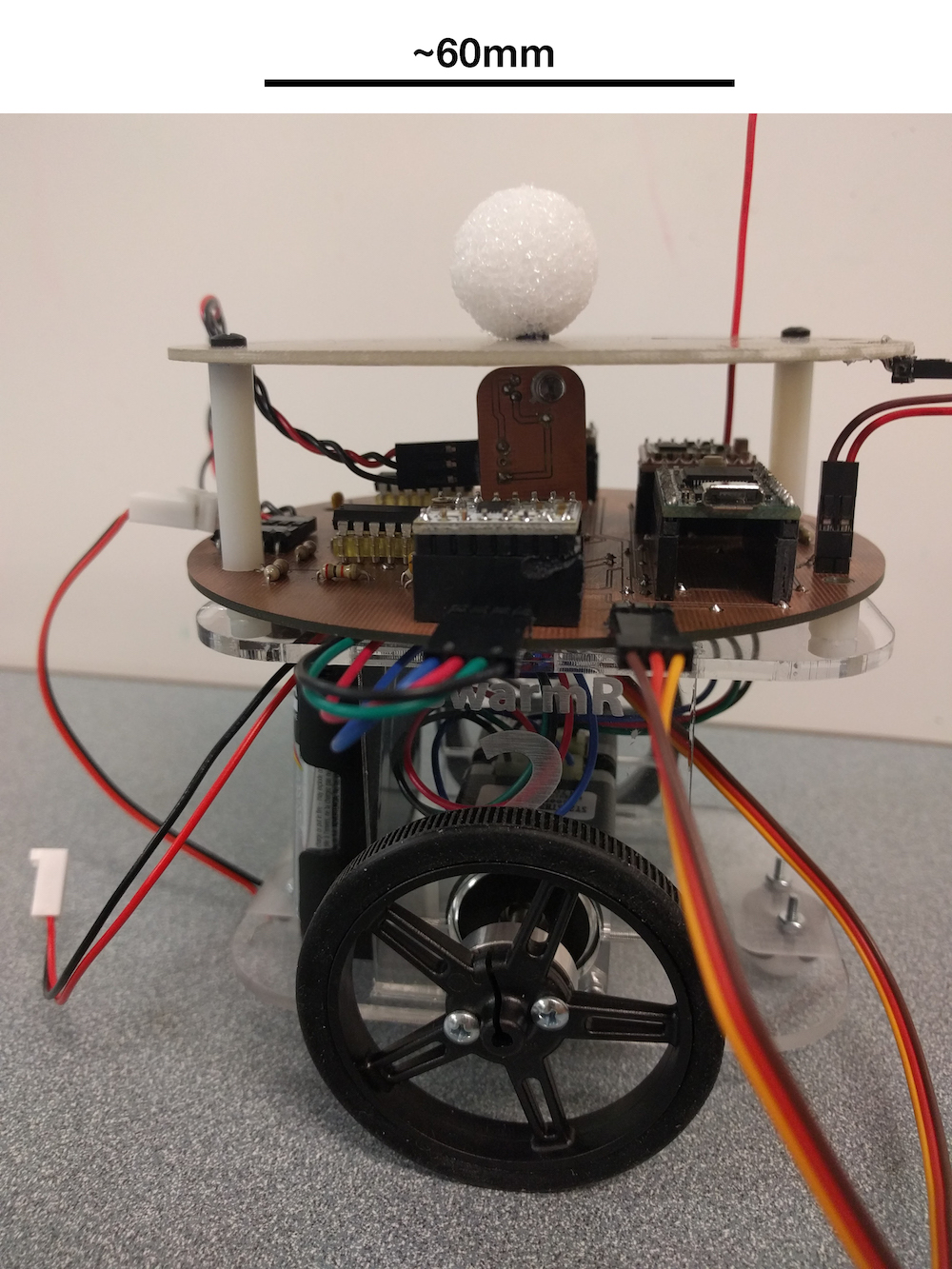}
    \caption{Photograph of a single robot.}
    \label{fig:robot_picture}
\end{figure}

The robot design focused on maximizing versatility while minimizing cost.  As depicted in Fig.~\ref{fig:hardware_block_diag}, the system consists of an \texttt{RFM69} radio, dual stepper motors and drivers, IR beacons and sensors, and a Teensy 3.2 ARM micro-controller.  Communication with the controlling computer is achieved with another \texttt{RFM69} connected through a micro-controller via USB to a computer.
Fig.~\ref{fig:robot_picture} depicts a single robot, employing two stepper motors driven with interrupt-controlled stepper drivers.  Two bearing balls are used as low-friction draggers on the front and back
for stability.

An IR LED on the top of the robot pulses at a frequency between 1-50~kHz (unique to each robot) to allow node identification.  The LED is covered with a white foam ball to evenly distribute the signal.  Two phototransistors are used as detectors, placed on either side of a vertically mounted PCB attached to a servo motor.  As the servo rotates back and forth, each phototransistor sweeps out a $180^\circ$ swath on each side of the robot. A custom analog circuit filters the raw phototransistor output, which is fed directly into one of the micro-controller's onboard ADCs.

\subsection{Sensor Signal Processing}
\begin{figure*}
    \centering
    \includegraphics[width=0.8\textwidth]{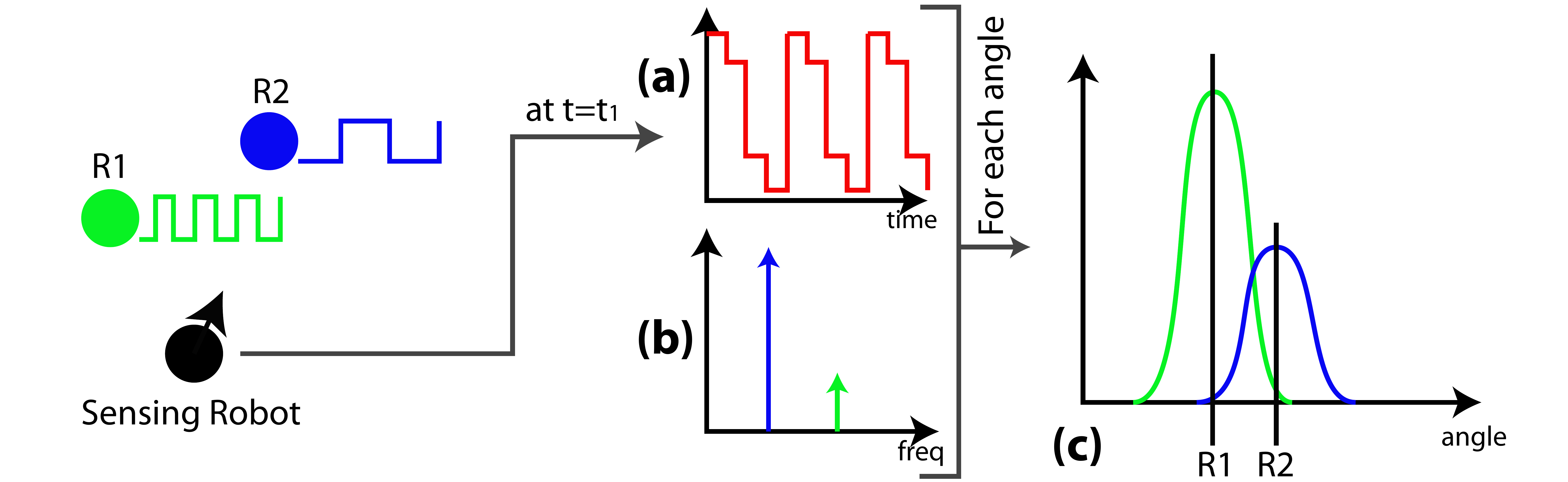}
    \caption{Data flow for converting raw sensor data to target range and bearing.}
    \label{fig:sensor_proc_process}
\end{figure*}

An FFT is performed on the incoming data at each sweep angle,\footnote{The Teensy audio library is used to perform the FFT, which uses 1024 samples at about 30kHz (standard audio sample rate).  This library has the benefit of using the DSP instructions available in the ARM core, so it is quite fast.} and because the IR LEDs are pulsed, the raw data consists of superimposed square waves of varying amplitude as depicted in Fig.~\ref{fig:sensor_proc_process}(a). Specific frequency bins correspond to each robot, and the values of only these bins are stored at each angle.

Once an entire $360^\circ$ sweep is obtained (from simultaneous $180^\circ$ sweeps), the magnitude of each bin at each angle can be plotted, as shown in Fig.~\ref{fig:sensor_proc_process}(c), where neighboring robots appear as signal peaks.  The frequency bin, angular location, and height of the peaks determine the nearby robot's identity, bearing, and range, respectively.

\begin{figure}
  \centering
  \includegraphics[width=0.4\textwidth]{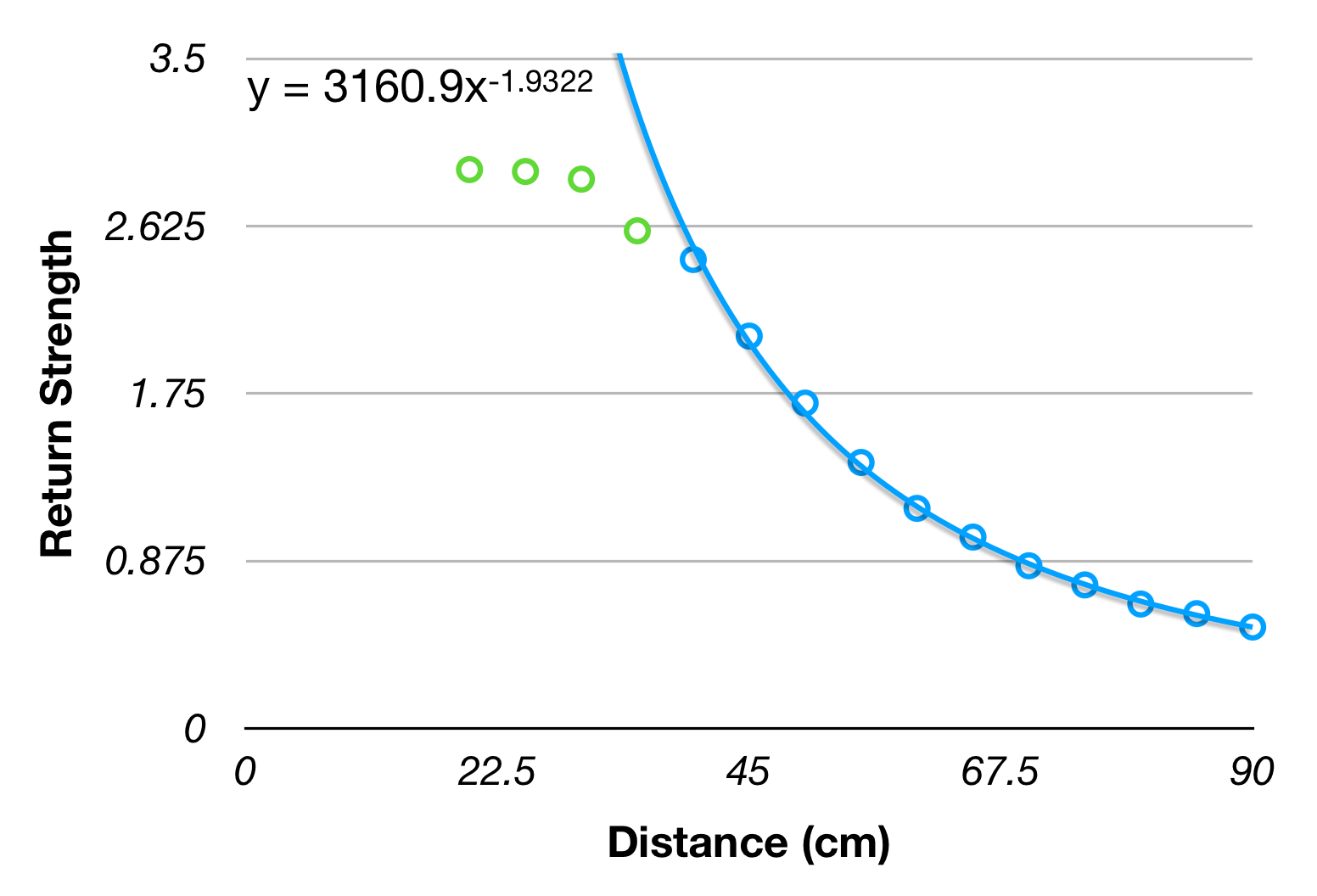}
  \caption{Plot showing the sensor signal magnitude for a target at varying distances, as well as the power-law best fit.}
  \label{fig:sensor_calib}
\end{figure}

The relationship between peak magnitude and distance can be modeled with a power law, given by
\begin{equation}
    D = {\text{constant}*\text{luminosity}\over{\text{surface area}}} = A M^{-2}
\end{equation}
where $M$ is the peak magnitude and $A$ is a constant.  In practice, there are significant variations between individual sensors and components used in the analog electronics, requiring each robot to be calibrated as shown in Fig.~\ref{fig:sensor_calib}.  A power-law fit is used to allow the exponent to vary in the event that anisotropies of the transmitter cause deviation from the ideal inverse square law. Note that the low-range readings were ignored for the purposes of fitting, since they deviate strongly from the power-law curve due to saturation and occlusion by the transmitter mount.

\subsection{Determination of Noise Matrices}
\begin{figure}
    \centering
    \includegraphics[width=0.4\textwidth]{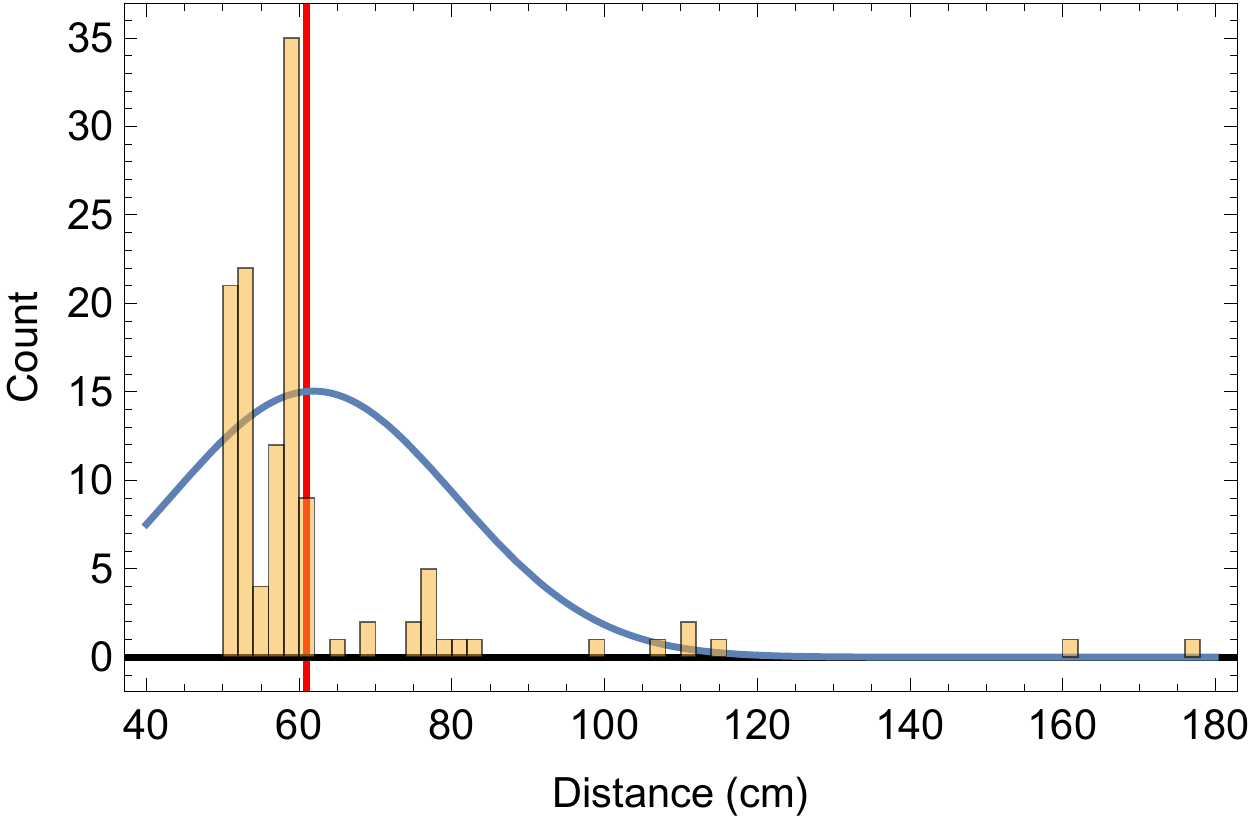}
    \caption{Histogram of distance measured between two robots, with actual distance shown in red.  A Gaussian distribution with the same mean and variance is shown in blue.}
    \label{fig:distance_histo}
\end{figure}

As described in Sec.~\ref{sec:kalman}, the Kalman filter requires the error from the motion model and sensors to be characterized.  Generally, we have chosen error standard deviations that are slightly higher than measured or estimated values, which may be pessimistic, but they have yielded good experimental results.
To quantify error in the sensor model ($\sigma_{\rm dist}$), two robots were placed 61~cm apart, and the estimated distance was recorded while one robot rotated in place.  Fig.~\ref{fig:distance_histo} shows a representative histogram from this procedure.

Although standard deviation of range sensing in this example is 18.6~cm, most of the readings are quite accurate, as indicated by a standard deviation of 7.2~cm when outliers (distances $>$ 90~cm) are removed.  The outliers are caused primarily by blind spots from the supporting standoffs that hold up the IR LED board.  Choosing to ignore outliers in computation of $\sigma_{\rm dist}$ results in a filter that generally tracks well but diverges when a highly erroneous value is received.  An alternate approach uses the actual skewed value of $\sigma_{\rm dist}$, giving a slower converging (but more stable) filter.  The latter approach was adopted herein with $\sigma_{\rm dist} = 15$~cm.  A possible more sophisticated approach would be to pre-filter sensor data and throw out readings which are vastly different from expected.

The variance in sensor angle is ideally only limited by the resolution of the angular scan, or $2^\circ$, but is practically higher due to robot movement and delay in the sensing method.  The update rate is approximately 1~Hz, giving a typical delay of 0.5~s.  Considering robot rotation speed to be $20^\circ/{\rm s}$ on average, the average error is $10^\circ$.  For the following experiments, a standard deviation of $\sigma_{\rm angle} = 0.15$~rad or $8^\circ$ was used.

The system propagation matrices required for the Kalman filter are more challenging to determine.  Typically, the model is very accurate because we are using steppers, but occasionally the robots slip on the floor for about 1~s.  The average error after a single predict step (run at a 30~Hz update rate) is found assuming a typical forward speed of 3~cm/s, giving an error of 0.1~cm as the wheels slip and suggesting the choice of $\sigma_{x}=\sigma_{y}=0.1$~cm.  Taking the typical rotation value of $20^\circ/{\rm s}$, the maximum change in angle after 1/30~s is 0.02 rad, and $\sigma_{\theta}=0.03$~rad was used.

%-----------------------------------------
\section{Localization Algorithms}
The sensing method and hardware described in Sec.~\ref{section:platform} yield range and bearing estimates of robots that are 1~m away or closer.  Individual robots report this information to a central controller, who must combine the estimates to obtain a global picture of the swarm.  This problem will first be solved using a direct estimation procedure that does not require the initial state of the swarm to be known.  Due to the computational complexity and limitations of the direct method, we later develop a more efficient tracking method based on Kalman filtering that more optimally deals with sensor error.

\subsection{Direct Estimation Method}
\label{sec:direct_est}
\begin{figure}
  \centering
  \includegraphics[width=0.2\textwidth]{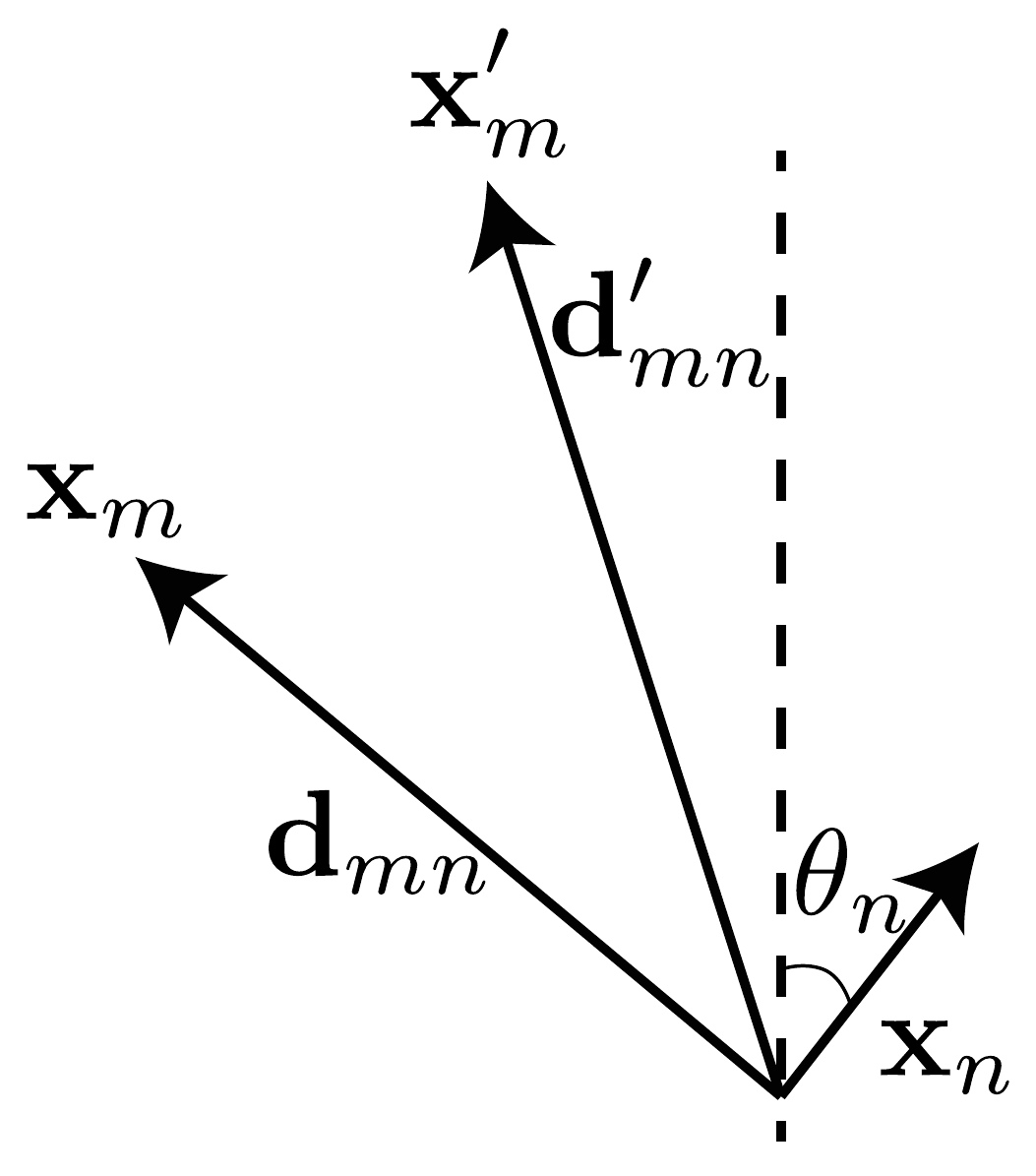}
  \caption{Geometry of Detected and Actual Location of Adjacent Robots}
  \label{figure:minimization_diag}
\end{figure}
We follow an approach similar to \cite{ahmad}, seeking to minimize the least-square error between the global robot map and reported local positions.  However, here we do not assume stationary landmarks nor do we exploit knowledge of robot velocities.  Consider the geometry shown in Figure \ref{figure:minimization_diag}, where $\xv_{p}$ and $\theta_{p}$ give the global estimated position and rotation of robot $p$, respectively, where $p \in \{m, n\}$ for the $m$th and $n$th robots.  The displacement vector $\dv_{mn}$ can be computed between the robots in the global model.  Likewise, robot $n$ obtains a local estimate of the displacement vector to robot $m$ as $\dv'_{mn}$ using local range and bearing ($\theta'_{mn}$) estimates, resulting in a perceived location of $\xv'_{m}$.

Given $N$ total robots and that robot $n$ sees the set of robots denoted ${\cal M}_n$, the goal of the estimation procedure is to find the $\dv_{mn}$ that minimize
\begin{equation}
    \label{eqn:objective_function}
    \phi = \sum_{n=1}^{N} \sum_{m \in {\cal M}_n} |\dv_{mn}-\dv'_{mn}(d_{mn}, \theta'_{mn})|^{2},
\end{equation}
which is non-trivial to solve for more than two robots.  In this work, minimization of $\phi$ was accomplished using the sequential least-squares programming minimization algorithm (SLSQP) from the Python \texttt{SciPy} library.  Special care was taken to program the objective function (\ref{eqn:objective_function}) as a matrix operation using \texttt{NumPy} to allow fast computation in Python.

Note that the minimization algorithm only uses sensor information on the \emph{relative} positions of the robots, as robots know nothing about the \emph{global} environment.  Therefore, the global rotation and translation of the complete swarm are free parameters.  For evaluation purposes, the semi-random values of these parameters as returned by the minimization algorithm are modified to optimally fit the estimated positions to the known positions.  This is accomplished
by aligning centroids and applying the Kabsch algorithm \cite{kabsch} to find the appropriate rotation.

\begin{figure*}
\centering
\includegraphics[width=0.8\textwidth]{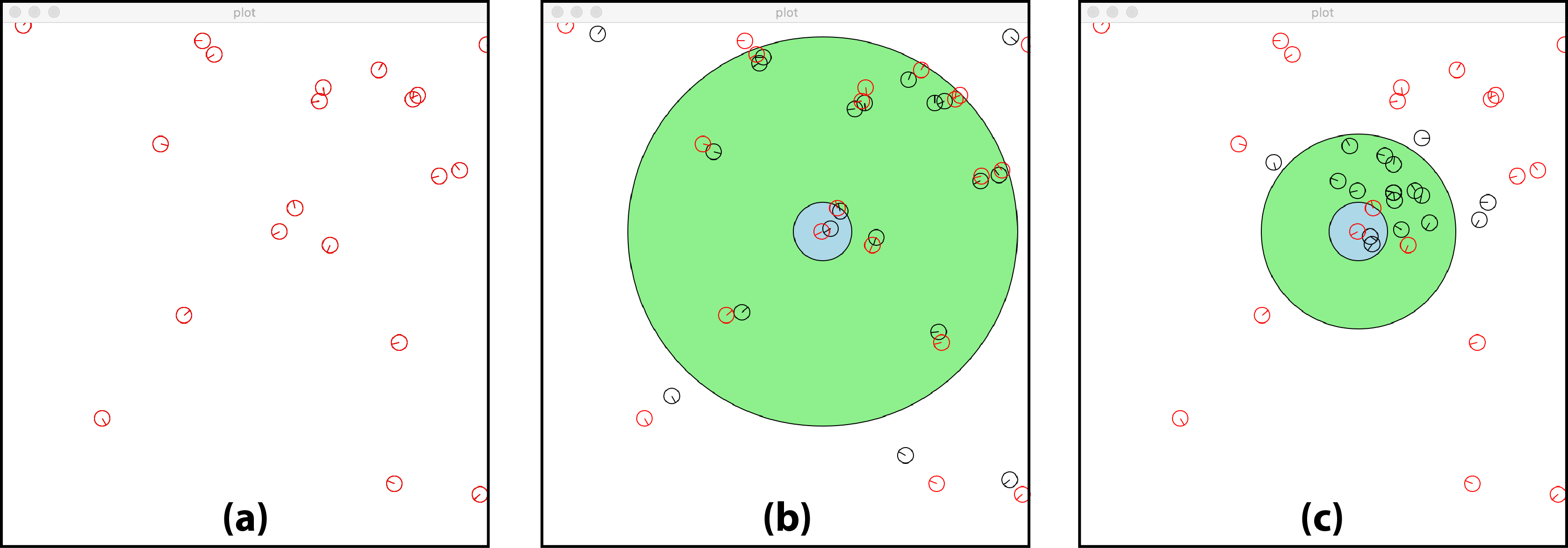}
\caption{Direct estimation algorithm simulation with (a) no error, (b) simulated noise, and (c) insufficient data.  The red circles are the actual positions, the black the calculated positions.}
\label{fig:min_all}
\end{figure*}

Fig.~\ref{fig:min_all}(a) shows the solution given by this minimization method for 20 robots, assuming that each robot can see all others and that there is no local estimation error.  Error in the global map is effectively zero, and this computation required roughly $0.5$~s.
Fig.~\ref{fig:min_all}(b) shows the effect of including moderate sensor limitations and estimation error.  Here, we introduced random Gaussian noise in distance measurements with a standard deviation given by the radius of the blue circle, Gaussian angular noise with a standard deviation of $6^\circ$, and a sensing range of half of the field size, given by the green circle.  Results deviate slightly from the exact solution, but average error is on the order of the robot size.  Computational time did not increase appreciably compared to the ideal case.  

Fig.~\ref{fig:min_all}(c) considers a case when the algorithm fails,  where the maximum sensing range is decreased further, resulting in an incorrect solution.  This occurs due to insufficient data about the robot adjacencies, since there are other incorrect solutions that may minimize error.  There are a total of $3N-3$ degrees of freedom (DOF), since each robot has three DOF and the arbitrary rotation and position of the swarm also has three DOF.  Each sensor reading provides two pieces of information (bearing and range).  We clearly require
\begin{equation}
  2 \sum _n |{\cal M}_n| \geq 3 N - 3,
\end{equation}
where $|\cdot|$ denotes set cardinality.  This requirement can be better understood by assuming each node sees $M$ neighbors, and in this case we require $2 N M \geq 3 N - 3$, or $M \geq 3(N-1)/(2N) \approx 3/2$ for large $N$.

\subsection{Kalman Filtering}
\label{sec:kalman}
The minimization strategy of the previous section works well for initialization, but is highly sensitive to sensor noise and can only update at the sensor refresh rate.  To solve these problems, we turn to Kalman Filters, using a method similar to the landmark localization problem described in \cite{labbe} and \cite{peter} where the landmarks are in fact mobile robots themselves, as described in \cite{martinelli}.

%\subsection{Algorithm Implementation} %\label{section:algorithmimplementation}
A distributed algorithm is desirable, since it would theoretically scale better for large swarms, but this would additionally require inter-robot communication.  Distributing the algorithm is partially supported by using separate Kalman filters to track the state of each robot.  However, the algorithm still uses the states of other robots in the update step, which is required to track the complete swarm. For our predict step, we implement a differential drive model as described in \cite{dudek}.  We assume independent variables, so the system noise covariance is simply a diagonal matrix of parameter variances.

We first develop the measurement function $\bf h$, which transforms from state space to measurement space.  Each other robot has its own filter generating its own state estimate, so we can treat these states as known landmarks, and then apply the same landmark measurement model used in \cite{negenborn}.

If we have a robot $n$ that sees robots $m_1, m_2, ... \in {\cal M}_n$, then
\begin{equation}
    \label{eqn:measurement}
    {\mathbf h} (n) = 
    \begin{bmatrix}
        |\mathbf{d}'_{m_1 n}|^2 \\
        \tan^{-1}(x(\mathbf{d}'_{m_1 n})/y(\mathbf{d}'_{m_1 n}) - \theta_n \\
        |\mathbf{d}'_{m_2 n}|^2 \\
        \tan^{-1}(x(\mathbf{d}'_{m_2 n})/y(\mathbf{d}'_{m_2 n}) - \theta_n \\
        \vdots
    \end{bmatrix}.
\end{equation}
where $x(\mathbf{d})$ and $y(\mathbf{d})$ simply extract the $x$ and $y$ components of the vector $\mathbf{d}$.
In practice we use the two-argument inverse tangent function (\texttt{atan2}) to guarantee that the resulting sign is correct.
As with the motion model, we also model the noise, in this case from the sensor.  Assuming independent variables, we have
$\Rm = {\rm diag}([\sigma_{\rm dist}^2, \sigma_{\rm angle}^2, \sigma_{\rm dist}^2, \sigma_{\rm angle}^2, ...])$, where ${\rm diag}(\cdot)$ forms a diagonal matrix from its vector argument.

In \cite{martinelli}, an extended Kalman filter (EKF) is used to handle the system nonlinearities.  We found, however, that an EKF tended to diverge when faced with highly nonlinear robot trajectories.  To handle these cases, we instead implement an unscented Kalman filter (UKF).  Where the EKF simply linearizes the model around an operating point, the UKF takes a set of ``sigma points'' which it passes through the function.  These sigma points are chosen using the method in \cite{vandermerwe}, with the choice of number of sigma points $n=3$, since we have 3 state variables, $\kappa = 3-n = 0$, $\beta = 2$, and $\alpha = .00001$.  These are the choices made by Labbe in his similar landmark localization problem in \cite{labbe}.  We then run these sigma points through the unscented transform to effectively fit a Gaussian distribution to the result.

To implement this filter, Labbe's \texttt{FilterPy} library was used,  only requiring small adjustments for this application.  For the residual calculation in the update step, a special function must be provided to properly calculate the difference of angles due to modulo angular arithmetic (e.g. $359^\circ - 1^\circ = -2^\circ$, not $358^\circ$). Furthermore, the number and identity of robots that any given robot can see at any time may, and most likely will, vary from sensor scan to scan.  When calculating the measurement residual, we therefore automatically set appropriate matrix entries in the residual to 0 for robots for which we have no sensor data.  To handle the varying size of $\bf{h}$ (the measurement matrix), modifications to the \texttt{FilterPy} library were made to automatically copy $\Rm$ to scale it to the appropriate size on each update step.

\begin{figure}
    \centering
    \includegraphics[width=0.4\textwidth]{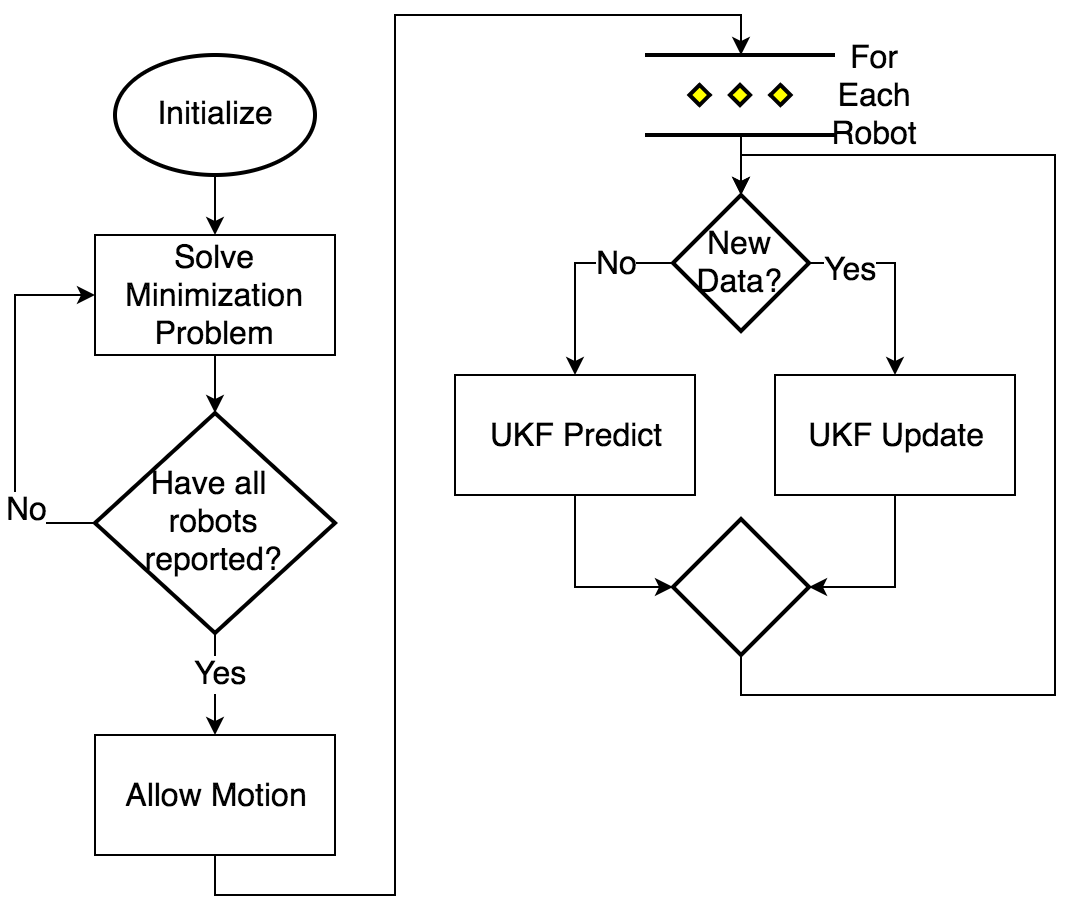}
    \caption{Flowchart showing the startup process for the UKF.  The direct estimation algorithm is used to determine the initial state before control is transferred to the UKF.}
    \label{fig:startup_process}
\end{figure}

The overall localization process is diagrammed in Fig.~\ref{fig:startup_process}.  Note that the direct estimation method Sec.~\ref{sec:direct_est} is used to determine the initial conditions, and control is then passed to the UKF.

%-----------------------------------------
\section{Testing}
\begin{figure}
  \centering
  \includegraphics[width=0.48\textwidth]{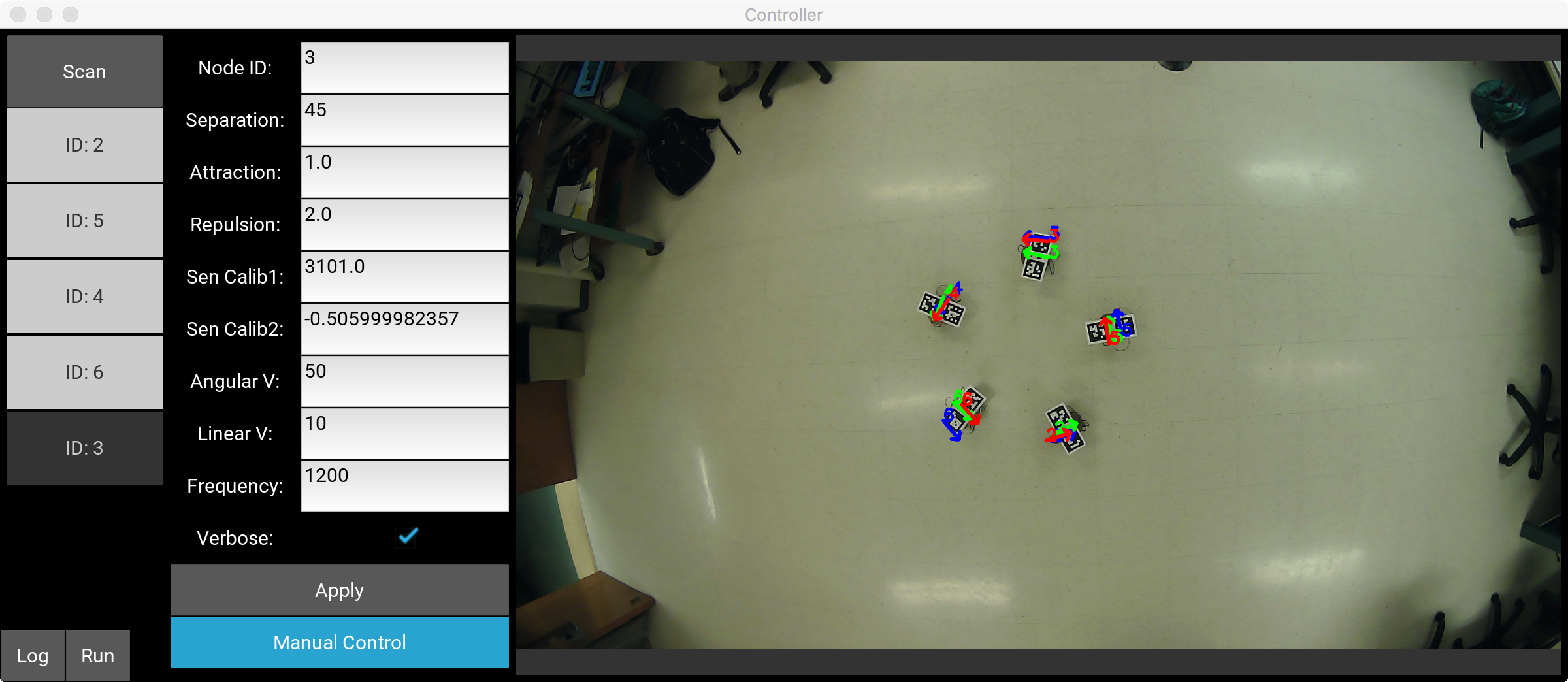}
  \caption{Image of the motion capture system running on 5 robots.      The different arrows correspond to the different localization systems: green for motion capture, red for minimization, blue for UKF.}
  \label{fig:mcap_screen}
\end{figure}

For comparison and testing purposes, each robot was equipped with ArUco tags \cite{garridojurado}, which were then processed through OpenCV to provide ground-truth position information.  The relative locations determined by the UKF were then translated and rotated to best fit to the known locations using the Kabsch algorithm \cite{kabsch}.  Simultaneously, the direct-estimation algorithm was run for comparison purposes.  The motion-capture system running is shown in Fig.~\ref{fig:mcap_screen}.

\begin{figure}
  \centering
  \includegraphics[width=0.5\textwidth]{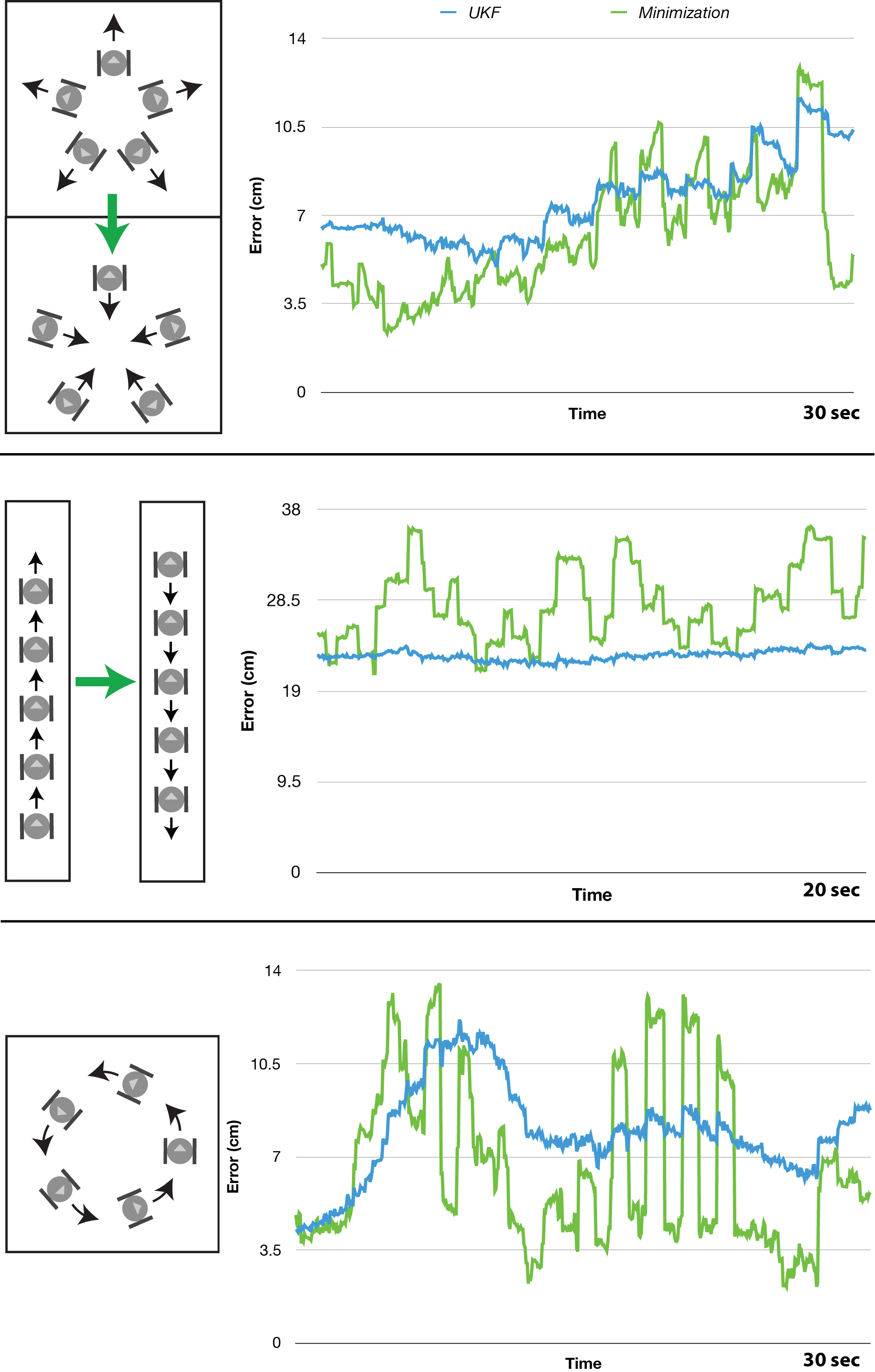}
  \caption{Plot of average error between calculated and actual       relative positions of robots over time for various 5-robot formations.}
  \label{fig:5_robots_all}
\end{figure}

Plots of the average error versus time for various maneuvers are shown in Fig.~\ref{fig:5_robots_all}, where the time scales are on the order of tens of seconds.  The primary source of error is periodic points of large range-error.  This can arise from a number of different sources, including blind spots created by support posts on the robots as well as cables becoming faulty on the
sensor boards due to constant rotation.  However, these effects do not cause the filter to diverge, and the overall average error remains about 10~cm, or on the order of the size of a robot.

The large amount of systematic error seen for the line formation is caused by accumulated systematic calibration error.  If each robot perceives its neighbors to be slightly further away than they actually are, then this error quickly accumulates in a line-like configuration.  This case highlights that relatively small systematic errors can quickly multiply into large problems in larger swarms.  A more sophisticated calibration process or general sensing system would be needed to mitigate this issue.

%-----------------------------------------
\section{Conclusion}
This paper has presented a low-cost multi-robot platform and accompanying algorithm based on the UKF that can estimate relative positions of robots in a small swarm.  The method was demonstrated for 5 robots, where the main limiting factors were sensor accuracy and robustness.  Accuracy of the position information was shown to be similar to that obtained with external capture systems, indicating that low-cost and deployable robot swarms are feasible. 

In order for the system to be practical, it is important for the operator to view not only the current relative configuration of the swarm, but also the configuration of the swarm in its current space.  This would make the system truly capable of mapping areas, which is a very promising application of swarms.  However, our robot nodes currently have no sensors capable of gathering information about their environment.  Care would be required to achieve this at low cost and without interfering with the existing sensing method.

Further work is also needed to develop a truly distributed version of our proposed UKF algorithm.  The current method relies on a centralized controller, but since each robot has a separate UKF, it should be feasible to distribute computation onto the robots themselves.  Another direction of future work could involve the design of a system with multiple robots employing reuse of the IR modulation frequencies, since this feature will be necessary to support of swarms of arbitrary size.

%Bibliography on a new page
\bibliography{swarmr} 
\bibliographystyle{IEEEtran}

\end{document}